\RequirePackage{fix-cm}
\documentclass[twocolumn]{svjour3}          % twocolumn
\smartqed  % flush right qed marks, e.g. at end of proof
\usepackage{graphicx}
\usepackage{natbib}
\usepackage{multirow}
\usepackage{amsmath}
\usepackage{amssymb}
\usepackage{bbm}
\usepackage{bm}
\usepackage{booktabs}
\usepackage{array} % for extended column definitions, e.g. >{\em}c
\usepackage{blindtext}
\usepackage{xspace}
\usepackage{url}

\usepackage{pifont}

\usepackage{xcolor}
\definecolor{citecolor}{HTML}{0071bc}
\usepackage[colorlinks,citecolor=citecolor]{hyperref}

\makeatletter
\renewcommand\paragraph{
  \@startsection{paragraph} % name
  {4} % level
  {\z@} % indent
  {.5em \@plus1ex \@minus.2ex} % beforeskip
  {-.5em} % afterskip
  {\normalfont\normalsize\bfseries} % style
}
\makeatother

\newcommand{\nickname}{3DTopia}
\newcommand{\etal}{\emph{et al.}\xspace}
\newcommand{\ie}{\emph{i.e.}\xspace}
\newcommand{\eg}{\emph{e.g.}\xspace}

\begin{document}
\sloppy

\title{3DTopia: Large Text-to-3D Generation Model \\ with Hybrid Diffusion Priors}

\author{
    Fangzhou Hong\textsuperscript{*} \and
    Jiaxiang Tang\textsuperscript{*} \and
    Ziang Cao\textsuperscript{*} \and
    Min Shi\textsuperscript{*} \and
    Tong Wu \and
    Zhaoxi Chen \and
    Shuai Yang \and
    Tengfei Wang \and
    Liang Pan \and
    Dahua Lin \and
    Ziwei Liu
}

\institute{
    Fangzhou Hong, Ziang Cao, Zhaoxi Chen, Ziwei Liu \at
    S-Lab, Nanyang Technological University, Singapore \\
    \email{\{fangzhou001, ziang.cao, zhaoxi001, ziwei.liu \}@ntu.edu.sg} \and
    Jiaxiang Tang \at
    National Key Lab of General AI, Peking University \\
    \email{tjx@pku.edu.cn} \and
    Min Shi \at
    Huazhong University of Science and Technology \\
    \email{min\_shi@hust.edu.cn} \and
    Tong Wu, Dahua Lin \at
    The Chinese University of Hong Kong \\
    \email{wt020@ie.cuhk.edu.hk} \and
    Liang Pan \at
    Shanghai Artificial Intelligence Laboratory \\
    \email{panliangde2007@gmail.com} \and
    Tengfei Wang, Shuai Yang, Tong Wu, Dahua Lin \at
    Shanghai Artificial Intelligence Laboratory \\
    \email{wangtengfei@pjlab.org.cn} \and
    \quad
    \textsuperscript{*}Equal Contribution
}

\date{Received: date / Accepted: date}

\maketitle

\begin{abstract}
    We present a two-stage text-to-3D generation system, namely \nickname{}, which generates high-quality general 3D assets within 5 minutes using hybrid diffusion priors. The first stage samples from a 3D diffusion prior directly learned from 3D data. Specifically, it is powered by a text-conditioned tri-plane latent diffusion model, which quickly generates coarse 3D samples for fast prototyping. The second stage utilizes 2D diffusion priors to further refine the texture of coarse 3D models from the first stage. The refinement consists of both latent and pixel space optimization for high-quality texture generation. To facilitate the training of the proposed system, we clean and caption the largest open-source 3D dataset, Objaverse, by combining the power of vision language models and large language models. Experiment results are reported qualitatively and quantitatively to show the performance of the proposed system. Our codes and models are available at \url{https://github.com/3DTopia/3DTopia}.
\end{abstract}

\section{Introduction}
Generating high-quality 3D assets from natural languages is a long quest in both industry and academia. Applications like games, visual effects, and virtual reality have a high demand for 3D assets. Production of high-quality 3D assets is time-consuming if made from scratch by human experts. Therefore, it is desirable to learn a model that generates 3D assets from natural languages. However, it is not trivial to train such a model. Compared with the amount of data used to train text-to-image models, the scale of existing text-3D paired dataset is too small. Moreover, 3D representations are normally more resource-consuming than 2D images, making the problem even more challenging.

\begin{figure}[t!]
    \centering
    \includegraphics[width=0.48\textwidth]{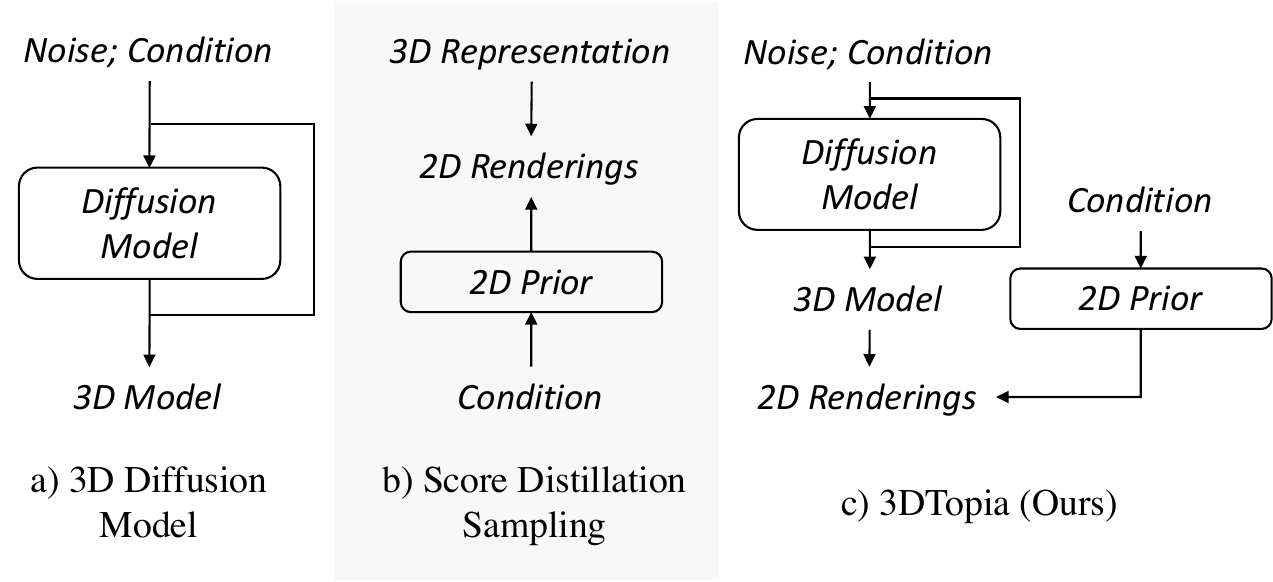}
    \vspace{-0.4cm}
    \caption{
    \textbf{Architecture Comparison of Different 3D Generation Paradigms.} We combine the advantages of feed-forward network and optimization-based methods and propose a two-stage generation system.
    }
    % \vspace{-0.2cm}
    \label{fig:arch_compare}
\end{figure}

There have been a lot of attempts in text-to-3D generation during the past years, which can be classified into two categories, as shown in Fig.~\ref{fig:arch_compare}. The first one trains a feed-forward diffusion network (Fig.~\ref{fig:arch_compare} a) from text-3D paired dataset, \eg, Point-E~\citep{pointe}, Shap-E~\citep{shapee}. Due to the feed-forward nature, the generation speed is fast, but the quality is inferior for limited dataset size. The second one uses Score Distillation Sampling~\citep{poole2022dreamfusion} (Fig.~\ref{fig:arch_compare} b) to optimize a 3D representation to match text description, \ie, DreamFusion~\citep{poole2022dreamfusion}. With the prior knowledge provided by powerful text-to-image models, the generation quality is high, but the optimization process is slow. To combine the best of both worlds, we proposed to use hybrid diffusion priors to form a two-stage system, namely \nickname{}. It can generate general 3D assets from user inputs with high quality and reasonable latency. The first stage uses a feed-forward network to ensure fast prototyping and iteration when the user is trying with different prompts. Once the user picks a coarse mesh from the first stage, our second stage would ensure a high-quality 3D asset with a few more minutes of optimization.

For the first stage, we learn a 3D diffusion prior directly from 3D data. To learn such a prior, the first but crucial decision to make is which 3D representation should we use. There have been a lot of NeRF-based 3D representations proposed, \ie, MLP~\citep{mildenhall2020nerf, wang2021neus}, voxel~\citep{yu_and_fridovichkeil2021plenoxels}, tri-plane~\citep{eg3d, chen2022tensorf}, point cloud~\citep{xu2022point}, hash tables~\citep{mueller2022instant}, Gaussian kernels~\citep{kerbl3Dgaussians}, \etal. Not all of them are suitable for the task of text-to-3D generation. We have two standards when choosing such a representation, \ie, a) efficiency in storage and compute; b) compatibility with neural networks. To achieve reasonable versatility, it is important to scale-up dataset size, which leads to pressure for data storage and conversion. To relieve the pressure, a compact 3D representation with reasonable expressiveness is preferable. Moreover, the data structure of this representation should be compatible with neural network operators since we are targeting a feed-forward network. Taking the above standards into consideration, we choose tri-plane~\citep{eg3d} as our 3D representation. A tri-plane consists of three axis-aligned 2D feature maps, making it both compact and easy to be processed by neural networks.

To further improve the efficiency of the tri-plane diffusion model learning, we adopt the idea of latent diffusion model (LDM)~\citep{ldm}. We first reconstruct tri-planes for each 3D asset in the dataset. Then they are encoded into the latent space by a tri-plane VAE. Then we learn to sample the tri-plane latent features by a diffusion model. Since we can flatten the tri-plane to a single 2D feature map, the diffusion model shares a similar structure with LDM~\citep{ldm}.

% \td{} second stage \textcolor{red}{@Jiaxiang}
For the second stage, we decide to take advantage of the powerful 2D diffusion prior to refine the coarse 3D samples from the first stage.
Specifically, we adopt the Score Distillation Sampling (SDS)~\citep{poole2022dreamfusion} for an optimization-driven refinement.
A well-established observation from previous works~\citep{shi2023mvdream,li2023sweetdreamer} is that SDS-based methods suffer from optimizing feasible geometry, but usually produce rich-detailed texture.
Given the coarse geometry generated by our first-stage feed-forward model, we find that it takes much less time to refine its texture using SDS.
Furthermore, we explore different 2D diffusion priors and design a hybrid refinement pipeline using both latent-space and pixel-space diffusion models~\citep{ldm,deepfloydif}.
In total, our refinement stage takes about 4 minutes to generate realistic and detailed texture on the coarse mesh from the first stage.

% \td{} caption and cleaning \textcolor{red}{@Shi Min} \textcolor{red}{@Wu Tong}

Recent text-to-image research has shown the importance of high-quality captions~\citep{dalle3}. Fine-tuning on a small set of carefully chosen high-quality data samples can also greatly improve the generation quality. Therefore, we propose an automatic 3D data caption and cleaning pipeline, which produces 360K captions and a high-quality subset of 135K 3D objects for Objaverse~\citep{deitke2023objaverse}.

We report both qualitative and quantitative results for each part of our system to validate our design choice. Noticeably, \nickname{} outperforms Point-E and Shap-E on text-to-3D generation. Our contributions are concluded as follows,

\textbf{1)} We propose a two-stage text-to-3D generation system, \nickname{}, using hybrid diffusion priors, that enables both fast prototyping and high-quality 3D generation.

\textbf{2)} As the first stage, we explore the tri-plane latent diffusion model and achieve reasonable quality and versatility.

\textbf{3)} An optimization-based mesh refiner is proposed as the second stage using hybrid latent-space and pixel-space 2D diffusion priors, producing rich-detailed texture within 4 minutes.

\textbf{4)} We propose a 3D captioning and cleaning pipeline and contribute a high-quality subset of Objaverse~\citep{deitke2023objaverse} with detailed captions.

\section{Related Work}

\subsection{3D Generation}
Numerous research efforts have been directed towards advancing 3D generative models. One direction of research~\citep{chen2019learning,pointflow,eg3d,get3d,pointe,shapee,lion,meshdiffusion,3dldm,nfd,diffrf,voxe,difftf} involves sampling 3D assets directly through a single generative model, while others~\citep{poole2022dreamfusion,lin2023magic3d,wang2023prolificdreamer,chen2023fantasia3d,wang2023score,tang2023dreamgaussian} opt for creating different 3D objects by mutually independent test-time optimization. Typically, the former approach offers higher efficiency, particularly when utilizing high-resolution 3D representations, while the latter excels in the fidelity of generation. In this paper, we aim to integrate the strengths of both strategies, constructing a comprehensive 3D generation method with the ability to efficiently produce high-quality 3D assets.%\textcolor{red}{@Ziang}

\subsubsection{Feed-forward Method}
Many popular 3D representations have been adopted in the feed-forward generative modeling literature. Point clouds~\citep{pointe,pointflow} are flexible and adaptive to various categories, but hardly represent the watertight and solid surface of most 3D objects. Besides, mesh-based~\citep{get3d,meshdiffusion,chen2023primdiffusion} and voxel-based methods~\citep{diffrf,voxe} are expressive for 3D assets. Relying on the explicit geometric prior, those two explicit methods are convenient for fast querying and evaluating. However, those explicit representations are hard to extend to high-resolution data due to high memory usage. Neural field-based methods~\citep{chen2019learning,3dldm,hong2022eva3d} encode the geometry and texture implicitly, which avoids the difficulties in high-resolution conditions but increases the inference time. Compared with those representations mentioned before, by introducing geometric relations into latent features, tri-plane representation can achieve a promising trade-off between efficiency and performance. Tri-plane-based methods~\citep{eg3d,wang2023rodin,nfd,difftf,xie2023citydreamer,hu2024structldm} can adopt high-resolution data to represent various 3D assets while reducing huge memory loading. Therefore, for the efficient generation of coarse 3D objects, we employ tri-plane as our representation.
% In conclusion, our proposed method, namely \nickname{}, can generate a coarse 3D object within 1 minute.%\textcolor{red}{@Ziang}

\subsubsection{Optimization-based Method}

Optimization-based approaches combine differentiable 3D representations~\citep{mildenhall2020nerf,kerbl3Dgaussians} with 2D priors~\citep{ddpm,ldm,saharia2022photorealistic} to achieve open-vocabulary 3D generation~\citep{jain2022zero,poole2022dreamfusion,wang2023score,mohammad2022clip,michel2022text2mesh,hong2022avatarclip}.
These approaches aim to optimize a 3D representation like NeRF~\citep{mildenhall2020nerf} such that images rendered from it resemble those produced from pretrained 2D diffusion models.
In this process, the rich information from 2D diffusion models is distilled into the trainable 3D representation, ensuring both realism and 3D consistency.
Further researches continue to improve generation quality with various enhancements~\citep{lin2023magic3d,tsalicoglou2023textmesh,zhu2023hifa,yu2023points,li2023focaldreamer,chen2023it3d,wang2023prolificdreamer,huang2023dreamtime,metzer2022latent,chen2023fantasia3d}, address the multi-face Janus problem~\citep{shi2023mvdream,li2023sweetdreamer,qiu2023richdreamer}, reduces optimization time~\citep{tang2023dreamgaussian,chen2023gsgen,yi2023gaussiandreamer}, and explore further applications~\citep{zhuang2023dreameditor,singer2023text,raj2023dreambooth3d}.
Although these methods still require comparably longer generation time to feed-forward method, the texture quality is usually better due to the powerful 2D diffusion prior.
Using a coarse mesh from our initial model, we have designed an optimization-based refinement stage that significantly improves texture quality in about 4 minutes.

\begin{figure*}[t!]
    \centering
    \includegraphics[width=\textwidth]{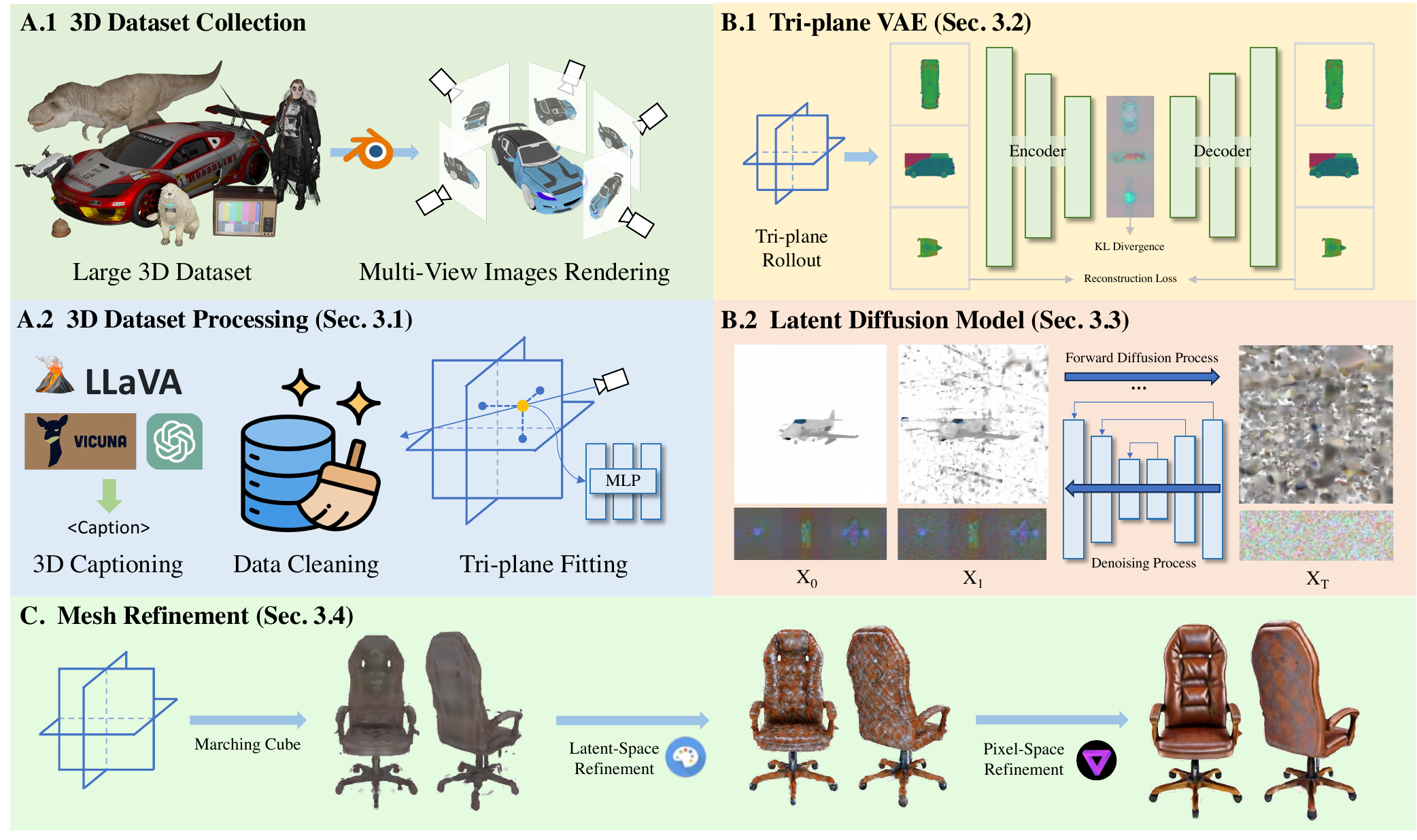}
    \vspace{-0.4cm}
    \caption{
    \textbf{Overview of our text-to-3D generation system.} We propose a two-stage generation system. The first system is a text-guided latent diffusion model (B.1 and B.2). To train the diffusion model, we first prepare a large text-3D paired dataset. We utilize a large web-crawled 3D dataset, Objaverse~\citep{deitke2023objaverse}, and render multi-view images from 3D assets (A.1). Then, multiple large language models are used for captioning and data cleaning. We choose to use tri-plane to parameterize 3D models (A.2). For the second stage, we use Score Distillation Sampling (SDS) for mesh refinement (C). It consists of two steps, \ie, latent-space refinement and pixel-space refinement.
    }
    % \vspace{-0.2cm}
    \label{fig:overview}
\end{figure*}

\subsection{Captioning}
% \textcolor{red}{@Shi Min}
Large-scale 3D object caption data is vital for multi-modal understanding and text-guided generation. Early works~\citep{text2shape,scan2cap,ulip} relying on 3D text data usually use small-scale human-annotated data for training, like ShapeNet~\citep{shapenet} and ScanNet~\citep{scannet}. To address the shortage of 3D data in multi-modal learning, ULIP-2~\citep{ulip2} first proposes to use a 2D caption model to generate 3D object descriptions at scale. They render the 3D object into 2D images from different views, and use 2D caption model~\citep{blip2} to generate captions automatically. Based on this render-and-caption pipeline, Cap3D~\citep{cap3d} systematically studied and improved the data curation pipeline. They generate more captions for each view and filter the low-quality ones with CLIP~\citep{clip}, and use GPT-4 to consolidate all the captions into one concise captions. Cap3D also demonstrates the necessity and effectiveness of high-quality 3D object annotations for 3D object generation model. Our annotation generation pipeline is similar to Cap3D, but with two major differences: 1) we leverage an advanced multi-modal large language model, LLaVA~\citep{llava} to generate more informative captions; 2): we integrate additional text processing to distill the essential information from the text, eliminating template-like responses and subjective judgments.

% ULIP, ULIP-2, multi-view rendering on shapenet -> captions -> train contrastive learning models
% Cap3D, multi-view images -> caption -> clip filter -> GPT-4 fusion
% Low quality due to less 'modern' image captioner; high cost on GPT-4 API calling; 
% we employ advanced VLM, add more intermediate text processing to obtain high-quality annotation.

\section{Methods}

The whole pipeline of \nickname{} is shown in Fig.~\ref{fig:overview}. We first perform 3D dataset collection and processing, including 3D captioning (Sec.~\ref{sec:3d_captioning}), data cleaning (Sec.~\ref{sec:data_cleaning}), and tri-plane fitting (Sec.~\ref{sec:triplane_fitting}). As the first stage, we train a tri-plane variational auto-encoder (VAE) (Sec.~\ref{sec:triplane_vae}). Based on it, a tri-plane latent diffusion model (LDM) is trained to sample coarse 3D models (Sec.~\ref{sec:triplane_ldm}). The second stage further refines the texture of generated 3D models (Sec.~\ref{sec:refinement}).

\subsection{Dataset Preparation}

\subsubsection{3D Captioning} \label{sec:3d_captioning}
\begin{figure*}[t!]
    \centering
    \includegraphics[width=\textwidth]{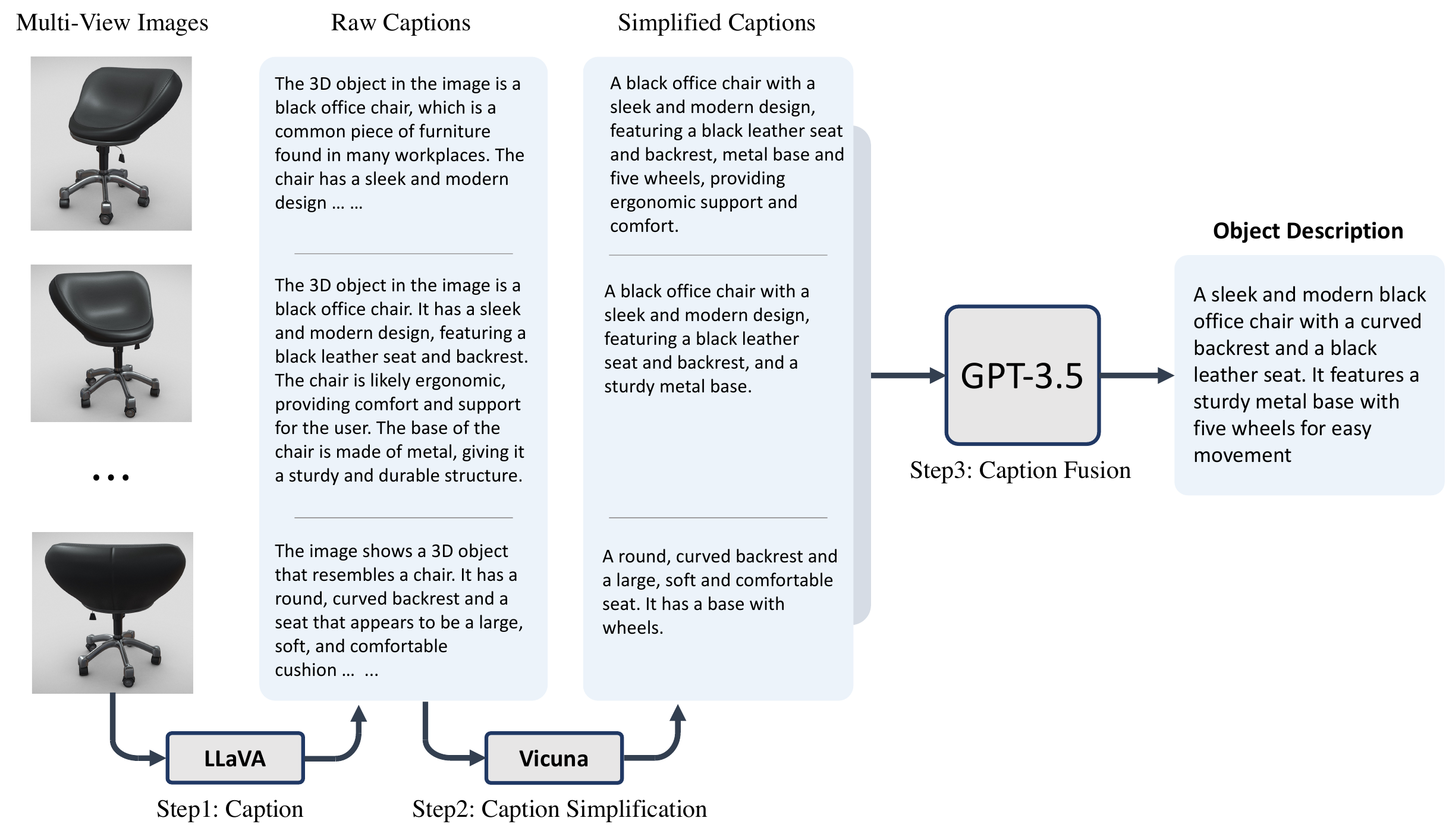}
    \caption{
        \textbf{3D captioning pipeline}. We first use LLaVA to generate raw captions for multi-view renderings, which is then simplified by Vicuna. Then we use GPT-3.5 to aggregate multi-view captions into a single caption.
    }
    \label{fig:caption-generation-pipeline}
\end{figure*}

To enhance the quality of training data, we construct an automated annotation pipeline to generate 360K captions for 3D objects. We refer to this dataset as 3DTopia-360K.

\begin{table}[b]
\caption{\textbf{Data statistics of 3DTopia-360K.}}
\label{tab:data-statistics}
\centering
\begin{tabular}{l|cc}
\toprule
             & \#samples & Avg. Length \\ \midrule
Cap3D~\citep{cap3d}      & 661,576   & 56.71      \\
3DTopia-360K & 361,357   & \textbf{132.13}     \\ \bottomrule
\end{tabular}
\end{table}

\textbf{Data annotation pipeline.}
Inspired by Cap3D~\citep{cap3d}, we generate captions for 3D objects by aggregating the descriptions from multiple views. This pipeline consists of three steps, as illustrated in Fig.~\ref{fig:caption-generation-pipeline}. Initially, a 2D image captioning model generates raw descriptions from multi-view images of a 3D object. Then, we prompt a language model to extract useful information from these raw descriptions to simplify them. Subsequently, a language model will consolidate all these captions into a single description for the 3D object. 
Different from Cap3D, we employ a multi-modal large language model, LLaVA~\citep{llava}, which can generate detailed description, as the captioning model. To effectively handle captions with long sequence lengths, we introduce additional language processing steps to improve the quality of annotations. Consequently, our captions encompass more details about the shape and texture of 3D objects. As shown in Table~\ref{tab:data-statistics}, the length of our captions exceeds those generated by Cap3D.

\textbf{Caption generation.}
We sampled 360K objects from Objaverse~\citep{deitke2023objaverse}\footnote{Available at \url{https://objaverse.allenai.org}} for annotation, filtering out objects without textures. Ten different views of each object are rendered against a white background. We employ LLaVA-13B to generate a caption for each rendered image using the following prompt:

\vspace{10pt}
{\centering
\quad
\setlength{\fboxsep}{4pt}
\fbox{
\parbox{65mm}{
\textbf{USER:} {\textcolor{red}{$<$image$>$}} I will show you a picture of a 3D object. Briefly describe the appearance and shape of it. \\
\textbf{Assistant}: ... ...}
}}
\vspace{10pt}

\textbf{Caption simplification.}
In this step, we use Vicuna-13B v1.5~\citep{vicuna2023}, an open-source large language model, to simplify the raw captions and extract useful information. Although LLaVA can generate relatively detailed descriptions, its output often contains many template-like cliché and subjective judgments, which are suitable for a concise caption. Additionally, we observe that the language model may struggle to consolidate excessively long captions. For example, the language model merely returns lists of captions rather than a cohesive summary. To simplify the caption, we utilize the following prompt:

\vspace{10pt}
{\centering
\quad
\setlength{\fboxsep}{4pt}
\fbox{
\parbox{65mm}{
\textbf{USER:} You will be given a description of an object. Please compress the description into one or two sentences. The details about the visual appearance and features must be retained. Please remove the irrelevant comments and contents that are not related to the object. Some examples are listed as follows: \\
\textit{\textcolor{gray}{8 examples … }} \\
\textbf{USER:} \textcolor{red}{$<$description$>$} \\
\textbf{ASSISTANT: }
}}}
\vspace{10pt}

\textbf{Caption fusion.}
After obtaining the simplified captions, we instruct GPT-3.5 to summarize all the captions into a unified description. To ensure the captions align with our desired format, we employ a few-shot strategy and furnish an example in the following prompt:

\vspace{10pt}
{\centering
\quad
\setlength{\fboxsep}{4pt}
\fbox{
\parbox{65mm}{
\textbf{USER:} Given a set of descriptions about the same 3D object, conclude these descriptions into one concise caption. The descriptions may contain contradictory information as each description comes from a certain view. In the output caption, keep the most specific information with more evidence and details. DO NOT generate ambiguous, contradictory or repeated information. Here is an example: \\
\textit{\textcolor{gray}{one example … }} \\
\textbf{USER:} \textcolor{red}{$<$list of descriptions$>$} \\
\textbf{ASSISTANT:}
}}}
\vspace{10pt}

In practice, we've observed that current open-source Large Language Model (LLM) struggles to execute this step effectively, as it necessitates complex reasoning and comparison processes. Consequently, we opt to utilize the GPT-3.5 API for improved quality.

\subsubsection{Data Cleaning} \label{sec:data_cleaning}
% \textcolor{red}{@Wu Tong}
Due to the wide range of data sources in the Objaverse~\citep{deitke2023objaverse} dataset, there is a significant variance in the content and quality of the data. We undertook a thorough data cleaning process before training. Specifically, our selection criteria focused on three aspects: \textit{content, geometry, and texture}. For content, we eliminated scene-level models, retaining only object-level models, which are more congruent with our framework. We also applied rigorous filters to both geometry and texture. A model is retained only if it demonstrates high precision and realism in either geometry or texture. This approach enabled us to discard a majority of artificially created objects of average quality, as well as flawed or incomplete scanned objects.

We ask human annotators to filter the dataset according to aforementioned criteria, resulting in a high-quality subset with 135K objects. We fine-tune our model on this high-quality subset after initial training on the full set to improve the generation quality.

\subsubsection{Tri-plane Fitting} \label{sec:triplane_fitting}

As an essential component of diffusion model training in Stage 1, we try to fit their tri-planes from various objects. Specifically, it consists of two steps: a) training the generalized shared decoder for various objects; and b) fitting the tri-planes of available objects. Specifically, tri-plane representation ($\mathbf{F}$) adopts three 2D spatial planes, $\mathbf{F}_{xy}$, $\mathbf{F}_{yz}$, and $\mathbf{F}_{xz}\in \mathbb{R}^{C \times W \times H}$, and a Multi-Layer Perceptrons (MLPs) decoder to map the corresponding latent feature to color (denoted as $\mathbf{c}=(r,g,b)$) and density (denoted as $\sigma$). Because of the mismatch between texture and geometry in distribution, adopting a unified decoder is hard to adapt to objects with complicated topology and texture. Therefore, different from the prior work~\citep{wang2023rodin,difftf}, we decouple the texture and geometry components and adopt two different Multi-Layer Perceptrons (MLPs) as decoders. For clarification, we denote the texture and density features as $\mathbf{F}^c$ and $\mathbf{F}^{\sigma}$. Thus, the texture features of the xy-plane can be represented by $\mathbf{F}^c_{xy}$. To achieve better performance of feature aggregation, we adopt channel-wise concatenation. The output of decoder, \ie, $\mathbf{c}$ and $\lambda$, can be formulated as
\begin{equation}
\begin{aligned}
    &\mathbf{c}=\mathrm{MLP}_{color}(\mathrm{Cat}(\mathbf{F}^c_{xy},\mathbf{F}^c_{yz},\mathbf{F}^c_{xz}))\\
    &\sigma=\mathrm{MLP}_{density}(\mathrm{Cat}(\mathbf{F}^{\sigma}_{xy},\mathbf{F}^{\sigma}_{yz},\mathbf{F}^{\sigma}_{xz}))
\end{aligned}
~.
\end{equation}

Note that to avoid the wide distribution of the values in tri-plane, we constrain the range of values from -5 to 5. Meanwhile, to reduce the noise in the fitted tri-plane, we add a strong regularization~\citep{nfd,difftf} during training, including TV loss and L1 loss. Together with L2 loss between ground truth ($G_c$) and predicted color, the shared decoder can be optimized by the joint loss function as follows:
\begin{equation}
\begin{aligned}
    &\mathcal{L}_{tri}=||G_c-\mathbf{c}||_2^2+\lambda_1\mathrm{TVloss}(\mathbf{F})+\lambda_2\||\mathbf{F}||
\end{aligned}
,
\end{equation}
where $\lambda_1$ and $\lambda_2$ are pre-defined coefficients to balance the different regularization.

By freezing the parameters of the trained shared decoder, we can efficiently fit tri-planes for all available objects, achieving a promising speed of \textless3 minutes per object.
% \textcolor{red}{@Ziang}

\subsection{Tri-plane VAE} \label{sec:triplane_vae}
Following the recent achievement in the diffusion-based generation model, we built our tri-plane generative model based on a latent diffusion model~\citep{ldm}. In comparison to the original tri-plane, the encoded feature focuses more on high-level semantic and abstract information. Besides, reducing in dimension of features accelerates the convergence of our diffusion model.

We build the VAE model based on the classic U-Net architecture. Inspired by RODIN~\citep{wang2023rodin}, we apply tri-plane rollout to flatten the 3D representation to a 2D feature map. To ease the learning of the diffusion model, we heavily reduce the latent dimension to as low as $32\times 32 \times 8 \times 3$. To ensure decent reconstruction quality, we set the weight of KL-divergence to $1\times 10^{-5}$. Moreover, as a regularization term, we apply Total Variance (TV) loss, with the weight of $2\times10^{-3}$, on the reconstructed tri-plane to reduce high-frequency artifacts.

Due to weak regularization of the tri-plane VAE training, the latent distribution is not guaranteed to be a normal distribution. Therefore, it is important to normalize the latent properly for better diffusion training. We calculate the mean $\mu$ and standard deviation $\sigma$ of the tri-plane latent over the whole dataset. Then we normalize the latent $l$ by $(l-\mu)/\sigma$.

% \textcolor{red}{@Fangzhou}

\subsection{Tri-plane Latent Diffusion} \label{sec:triplane_ldm}
Based on the tri-plane latent encoded by the Tri-plane VAE, we learn a diffusion model to sample over the distribution of 3D objects parameterized by Tri-plane. Diffusion model~\citep{ddpm} performs the generation process by a fixed Markov chain whose number of time steps is $T$. It learns the conditional probability distribution of data given the noisy version of data. By iteratively applying denoising steps, the model can recover the true data from the Gaussian noise.

\textbf{Diffusion process.} The diffusion process aims to gradually corrupt the distribution of the 3D object $f_0$ that we want by adding the noise iteratively as: 
\begin{equation}
\begin{aligned}
q(f_t|f_{t-1})=\mathcal{N}(f_t;\sqrt{1-\beta_t}f_{t-1};\beta_t \mathbf{I})
\end{aligned}
~,
\end{equation}
where $f_T,f_{T-1},...,f_0$ represent the latent features with different noised levels while $\beta_t$ and $\mathbf{I}$ refer to the forward process coefficient in step $t$ and unit matrix. Thus, we can derive the equation between $f_{t}$ and $f_0$ as: 
\begin{equation}
\begin{aligned}
q(f_t|f_0)=\mathcal{N}(f_t;\sqrt{\overline{\alpha}_t}f_0;(1-\overline{\alpha}_t) \mathbf{I})
\end{aligned}
~,
\end{equation}
where $\alpha=1-\beta$ and $\overline{\alpha}_t=\prod^t_{s=1}\alpha_s$. Based on it, we can retrieve latent features in arbitrary noised steps.

\textbf{Denoising process.} In contrast to the diffusion process, the denoising process aims to recover latent features from Gaussian noise, \ie, obtain the reversed transition probabilities $p_{\theta}(f_{t-1}|f_t)$. By sampling $f_T$ from a standard noise and introducing Bayes\textsc{\char13} theorem and specific parameterization, we can represent $p_{\theta}(f_{t-1}|f_t)$ as follows:
\begin{equation}
\begin{aligned}
    &p_{\theta}(f_{t-1}|f_t)=\mathcal{N}(f_{t-1};\mu_{\theta}(f_t,t),\Sigma_t)\\
    &\mu_{\theta}(f_t,t)=\dfrac{1}{\sqrt{\alpha_t}}(f_t-\dfrac{\beta_t}{\sqrt{1-\overline{\alpha}_t}}\epsilon_{\theta}(f_t,t))
\end{aligned}
~,
\end{equation}
where $\mu_{\theta}(f_t,t)$ is the noise added to $f_{t-1}$. Additionally, for simplification, $\Sigma_t$ adopts the pre-defined value as $\Sigma_{\theta}=\sigma^2_t\mathbf{I}$, where $\sigma^2_t=\dfrac{1-\overline{\alpha}_{t-1}}{1-\overline{\alpha}_t}\beta_t$. In conclusion, the objective of Stage-1 is to minimize the loss function as: 
\begin{equation}
\begin{aligned}
    &\mathcal{L}_{diff}=\mathbb{E}_{t,x_0,\epsilon}\left[||\epsilon-\epsilon_\theta(x_t,t)||^2\right]
\end{aligned}
,
\end{equation}

% \textcolor{red}{@Ziang}
% \td{} noise schedule: linear schedule + shift; CFG; DDIM
Inspired by recent studies in high-quality diffusion model training~\citep{hoogeboom2023simple}, it is important to scale noise schedules in accordance with different resolutions of input images. Therefore, we apply the SNR shift on top of a classic linear schedule as defined below:
\begin{equation}
\begin{aligned}
    &\text{SNR}_{\text{linear}}(t) = \overline{\alpha_t} / (1 - \overline{\alpha_t}) \\
    &\text{SNR}(t) = \text{SNR}_{\text{linear}}(t) / s^2
\end{aligned}
~,
\end{equation}

Moreover, in order to use classifier-free diffusion guidance~\citep{ho2022classifier}, the diffusion model is trained with $10\%$ of probability of zero condition.

\begin{figure*}[t!]
    \centering
    \includegraphics[width=\textwidth]{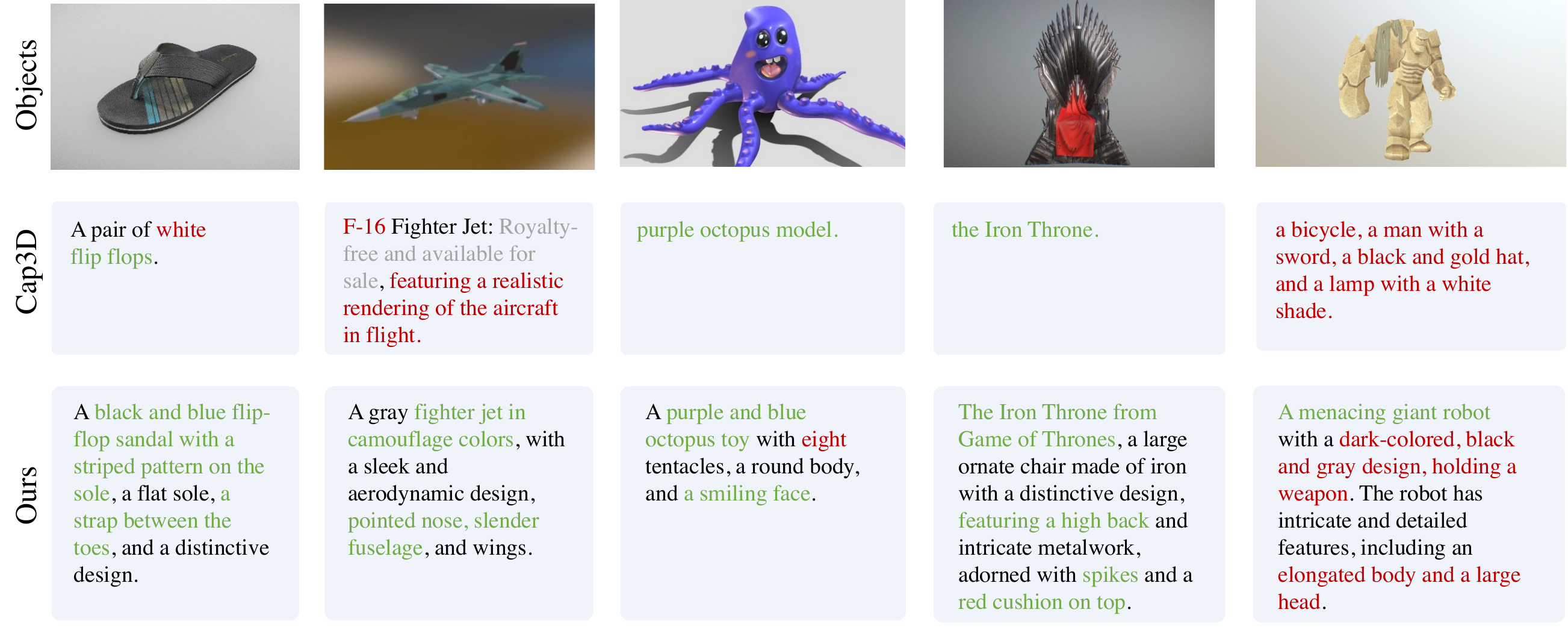}
    \caption{
    \textbf{Examples of our 3D captions and comparison with Cap3D}. Red texts are wrong parts. Green texts are correct parts. Compared with Cap3D captions, we provide longer captions with more details.
    }
    % \vspace{-0.2cm}
    \label{fig:caption-comparison}
\end{figure*}

\subsection{Optimization as Refinement} \label{sec:refinement}
SDS was initially introduced by DreamFusion~\citep{poole2022dreamfusion}, providing a framework that leverages pretrained 2D diffusion models as priors to optimize a parametric image generator. 
In our case, we take the exported polygonal mesh from the first stage as the input and aim to further refine the geometry and texture with SDS.

We first convert the mesh into a more optimization-friendly representation.
Specifically, we sample the signed distance function (SDF) within the bounding box of the mesh and construct an SDF grid, which can be easily converted back to mesh using marching cubes~\citep{lorensen1998marching}.
During this process, we also clean the first stage mesh by removing disconnected floater-like artifacts~\citep{tang2022nerf2mesh}.
Following~\citep{wei2023neumanifold}, we use a differentiable implementation of marching cubes, which allows gradient to be propagated to the SDF grid and an offset grid for geometry optimization.
For the appearance modeling, we use a hash gird~\citep{mueller2022instant} to encode it as 3D texture.
The converted mesh with hash grid texture can then be rendered using differentiable rasterization~\citep{Laine2020diffrast} to produce RGB images from arbitrary camera poses:
\begin{equation}
\mathbf{x} = g_\Theta(p),
\end{equation}
where $\mathbf{x}$ represents the rendered RGB image from the camera pose $p$, and $g_\Theta(\cdot)$ denotes the differentiable rendering function with optimizable parameters $\Theta$. 
To take advantage of the initial appearance from the first stage mesh, we optimize the hash grid texture for $512$ iterations using mean square error loss:
\begin{equation}
    \mathcal L = || \mathbf{x} - \mathbf{x}_\text{stage1} ||^2 ~,
\end{equation}
where $\mathbf{x}_\text{stage1}$ is the rendered RGB image from the same camera pose using the first stage mesh.

After the above pre-processing, we can apply SDS to refine this 3D representation.
We combine two types of 2D diffusion priors, namely, latent-space diffusion models and pixel-space diffusion models, for both robust and efficient refinement.

\subsubsection{Latent-space Refinement}
Latent diffusion models~\citep{ldm} adopts a VAE that compresses images into latent space:
\begin{equation}
    \mathbf{z} = \text{Enc}(\mathbf{x}) ~,
\end{equation}
where $\mathbf{z}$ is the latent code with a smaller resolution to save computation.
The SDS formulation for latent-space refinement can be expressed by:
\begin{equation}
\nabla_{\Theta} \mathcal{L}_\text{SDS}^\text{latent} = \mathbb{E}_{t, p, \mathbf{\epsilon}}
\left[
w(t)(\epsilon_\phi(\mathbf{z}; t, e) - \epsilon)
\frac {\partial \mathbf{z}} {\partial {\mathbf{x}}}
\frac {\partial \mathbf{x}} {\partial {\Theta}}
\right]~,
\end{equation}
where $t \sim \mathcal U(0.02, 0.98)$ is a randomly sampled timestep, $p$ is a randomly sampled camera pose orbiting the object center, $\mathbf{\epsilon} \sim \mathcal N (0, 1)$ is a random Gaussian noise, $w(t) = \sigma_t^2$ is a weighting function from DDPM~\citep{ddpm}, $\epsilon_\phi(\cdot)$ is the noise predicting function with a pretrained diffusion prior $\phi$, and $e$ is the text embedding.
By optimizing this objective, the denoising gradient $(\epsilon_\phi(\mathbf{z}; t, e) - \epsilon)$ that contains the guidance information is back-propagated to the latent code $\mathbf{z}$, which will be further back-propagated to the rendered image $\mathbf{x}$ and underlying parameters $\Theta$ of the 3D representation.
Although latent diffusion models produce high-resolution 2D images with rich details, it's observed that the VAE term $\frac {\partial \mathbf{z}} {\partial {\mathbf{x}}}$ leads to unstable gradient values and produce undesirable noisy artifacts, as shown in Fig.~\ref{fig:overview}.
Therefore, we only use the latent-space refinement to increase the diversity of the texture.

\begin{figure*}[t!]
    \centering
    \includegraphics[width=\textwidth]{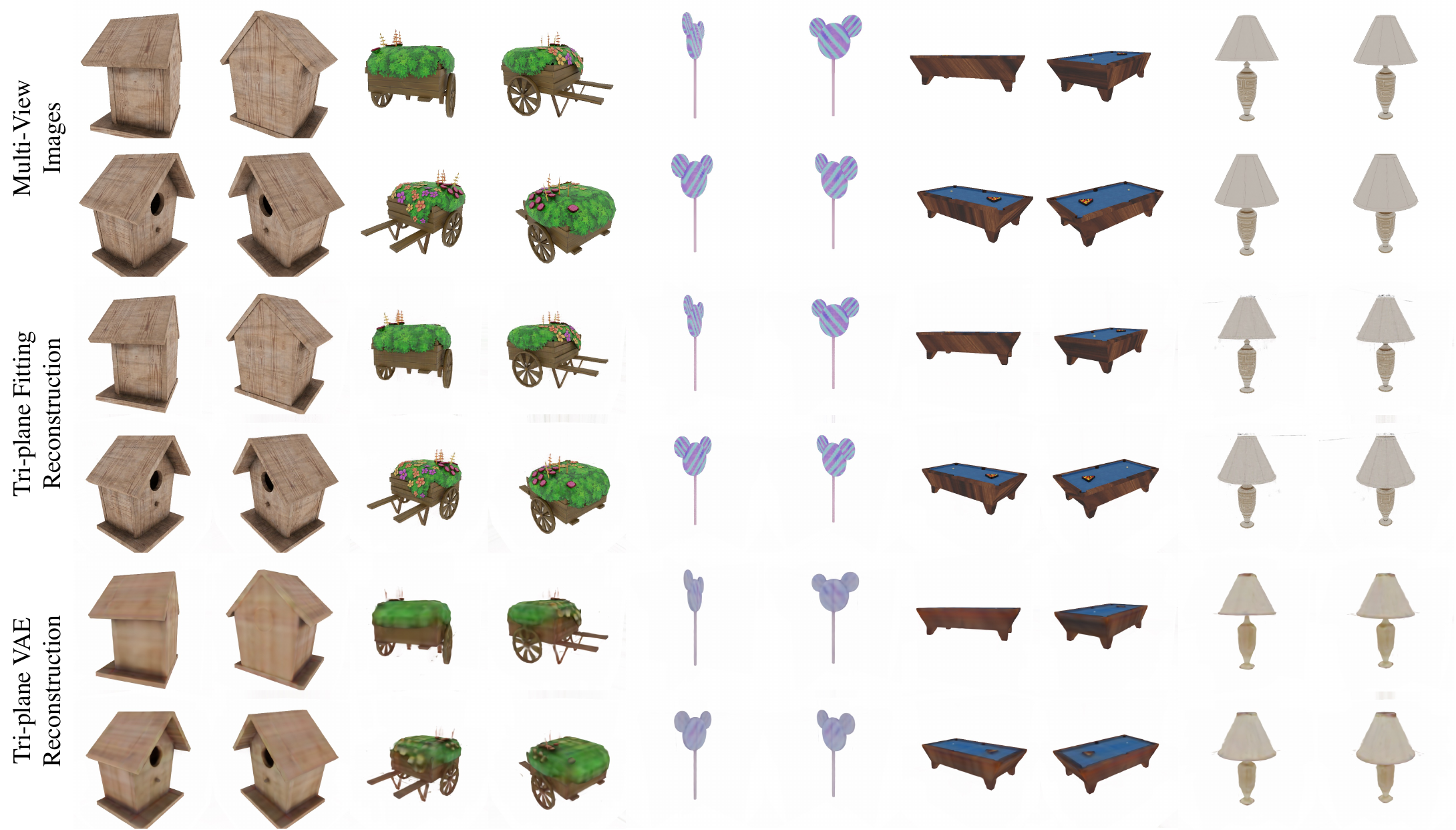}
    \caption{
    \textbf{Tri-plane fitting and VAE results}. We show ground truth multi-view images in the first row. The second row is the tri-plane fitting renderings. The third row shows the tri-plane VAE reconstruction results. We achieve a high compression rate while maintaining decent reconstruction quality.
    }
    \label{fig:main-recon}
\end{figure*}

\subsubsection{Pixel-space Refinement}
In contrast, pixel-space diffusion models~\citep{saharia2022photorealistic,deepfloydif} directly operate on the images.
Since we have the coarse texture from the latent-space refinement, we perform pixel-space refinement using a super-resolution diffusion prior.
Deepfloyd IF {\MakeUppercase{\romannumeral 2}}~\citep{deepfloydif} is the most suitable and available 2D diffusion model, which takes a low-resolution image as input and performs diffusion with higher resolution.
The SDS formulation can be represented by:
\begin{equation}
\nabla_{\Theta} \mathcal{L}_\text{SDS}^\text{pixel} = \mathbb{E}_{t, p, \mathbf{\epsilon}}
\left[
w(t)(\epsilon_\phi(\mathbf{x}; \mathbf{x}_\text{coarse}, t, e) - \epsilon)
\frac {\partial \mathbf{x}} {\partial {\Theta}}
\right]~,
\end{equation}
where $\mathbf{x}_\text{coarse}$ is the rendered image using the coarse texture.
As shown in Fig.~\ref{fig:overview}, we find that pixel-space refinement tends to produce smooth texture which fixes the noisy artifacts from the coarse texture, enriching the details while preserving the original color and style.

We can sequentially apply both latent-space and pixel-space refinement for the input mesh, which generates diverse and high-quality textures.
In practice, it's also possible to omit the latent-space refinement if the input mesh already has a favorable coarse texture.

\section{Experiments}

\subsection{Implementation Details}

The tri-plane dimension is $256\times256\times32\times3$ to ensure reconstruction quality. The tri-plane latent has a shape of $32\times32\times8\times3$. We first train the tri-plane VAE for 48k steps with a total batch size of 448, 112 NVIDIA A100 80GB GPUs, per-device batch size of 4 and learning rate of $1\times10^{-5}$. For the latent diffusion model, we first train it with 320K 3D objects for 118k steps with a total batch size of 512, 64 NVIDIA A100 80GB GPUs, per-device batch size of 8 and learning rate of $5\times10^{-5}$. Then we fine-tune the model on the 70k high-quality subset for 26K steps with the same configuration.

For all the results shown below, we use 200 steps DDIM~\citep{song2020denoising} and CFG sampling~\citep{ho2022classifier}, with the strength of $7.5$. 
For the second stage optimization, we append `\textit{best quality, extremely detailed, masterpiece, high resolution, high quality}' to the positive prompt and use `\textit{blur, lowres, cropped, low quality, worst quality, ugly, dark, shadow, oversaturated}' as the negative prompt.
For oriented objects, we also apply directional text prompts based on view directions following previous works~\citep{poole2022dreamfusion}.
The latent space refinement is trained for $800$ iterations, and the pixel space refinement is trained for $400$ iterations.
The learning rate is set to $0.01$ for the hash grid texture encoder, $0.001$ for MLPs, and $10^{-4}$ for the SDF and deformation grid.
We use a fixed rendering resolution at $512\times 512$, camera radius of $2.5$, and field of view angle of $49.1$ degree.
For the differentiable marching cubes resolution, we choose $128^3$. The final output mesh is remeshed and decimated to a maximum face number of $5\times 10^4$. We apply vertices offset loss and Laplacian smoothness loss to enforce smoothness of the mesh following~\citep{tang2022nerf2mesh}.

\begin{figure*}[t!]
    \centering
    \includegraphics[width=\textwidth]{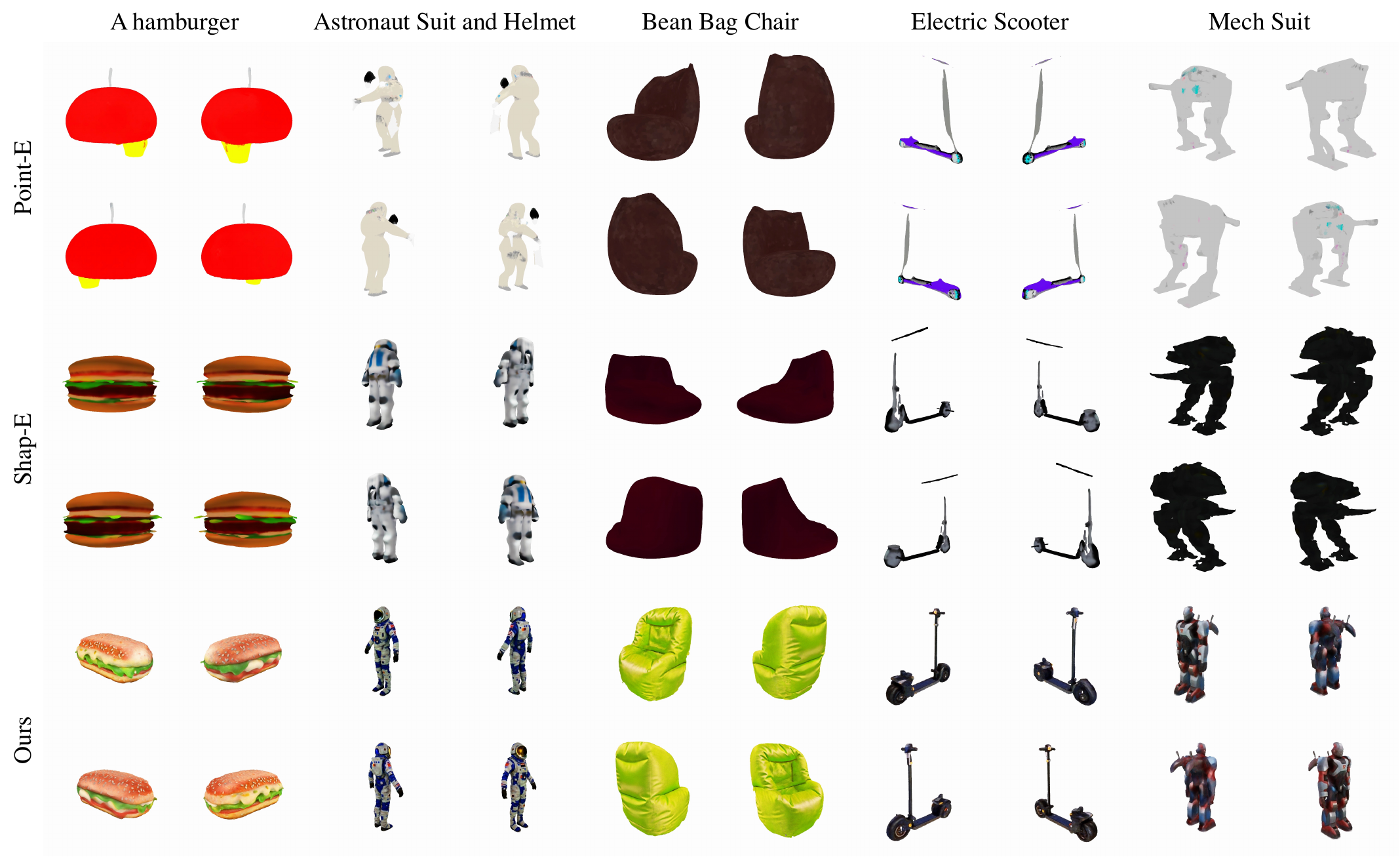}
    \caption{
    \textbf{Visual results of text-to-3D generation with \nickname{} and baseline methods}. The top row gives text prompts used for generation shown in each column. From top to bottom, we show generation results from Point-E, Shap-E and \nickname{}.
    }
    \label{fig:main-comparison}
\end{figure*}

\begin{figure*}[t!]
    \centering
    \includegraphics[width=\textwidth]{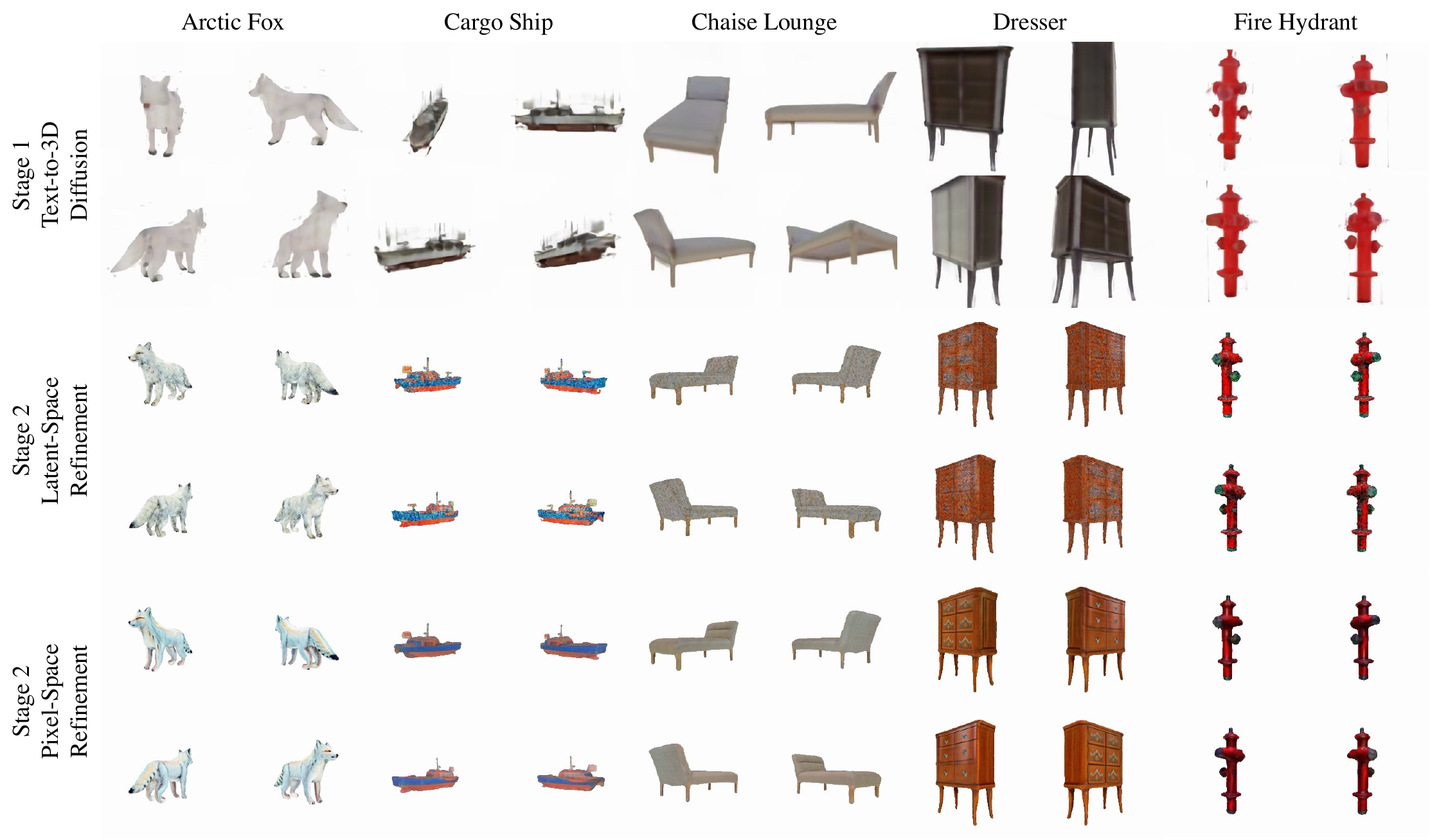}
    \caption{
    \textbf{Intermediate results of \nickname text-to-3D generation pipeline}. The top row gives text prompts used for generation. From top to bottom, we show generation results from stage 1 text-to-3D diffusion model, stage 2 latent-space refinement and stage 2 pixel-space refinement.
    }
    \label{fig:main-generation-process}
\end{figure*}

\subsection{Qualitative Results}

In this section, we show the qualitative results of different modules in \nickname{}, including 3D captioning, tri-plane fitting, tri-plane VAE, tri-plane diffusion and mesh refinement. We also show visual comparison of text-to-3D generation results with two baseline methods.

% \td{} Caption samples
As in Fig.~\ref{fig:caption-comparison}, we present some samples of our 3D captions. Compared with Cap3D, our descriptions encompass more details and concepts. For instance, our captions incorporate additional details that Cap3D overlooks. These include geometric shapes like ``pointed nose" and ``spikes", textures such as ``striped pattern" and ``camouflage colors," and poses like ``standing on its hind legs."
Meanwhile, there remains much room and potential for 3D caption generation. We observe that both Cap3D and our 3DTopia-360k datasets contain some noise, such as unreliable details due to hallucinations. This issue could potentially be mitigated by leveraging stronger vision-language models like GPT-4, which we defer to future work.

% \td{} Multi-view, triplane fitting, triplane VAE
As shown in Fig.~\ref{fig:main-recon}, we show the tri-plane fitting and tri-plane VAE results, which are rendered in four different views. As shown in the second row, the tri-plane fitting stage produces high-quality 3D parameterization, which lies a solid foundation for 3D generation training. The third row shows the reconstructed results of the tri-plane VAE. We achieve a high compression rate ($256\times$) while maintaining decent reconstruction quality of geometry. And we leave the high-quality texture generation to the second stage.

% \td{} Major figure: comparison with point-e shap-e
We conduct visual comparison with two baseline methods, Point-E~\citep{pointe} and Shap-E~\citep{shapee} in Fig.~\ref{fig:main-comparison}. Thanks to the two-stage generation system, \nickname{} produces high-quality generation results in terms of geometry and texture details. Compared with both baseline methods, the first stage of \nickname{} produces complete and correct geometry, especially in challenging cases of thin structure, \eg, electric scooter. Taking advantage of strong 2D priors in text-to-image models, the second stage of \nickname{} produces textures with higher fidelity and more details, \eg, wrinkles of bean bag chair.

% \td{} first stage results + second stage step 1 + second stage step 2
To better demonstrate the generation process of \nickname{}, we show intermediate results produced by different stages in Fig.~\ref{fig:main-generation-process}. The first row is the coarse generation results directly sampled from latent diffusion model. Our tri-plane latent diffusion model quickly produces 3D samples with correct geometry and coarse texture. The first step of stage 2 is the latent-space refinement, the results of which are shown in the second row. The latent-space refinement adds more high-frequency texture details on top of the first-stage geometry. The third row shows the results of the complete pipeline of \nickname{}. The pixel-space refinement further improves texture details and quality.

\begin{figure}[t!]
    \centering
    \includegraphics[width=0.5\textwidth]{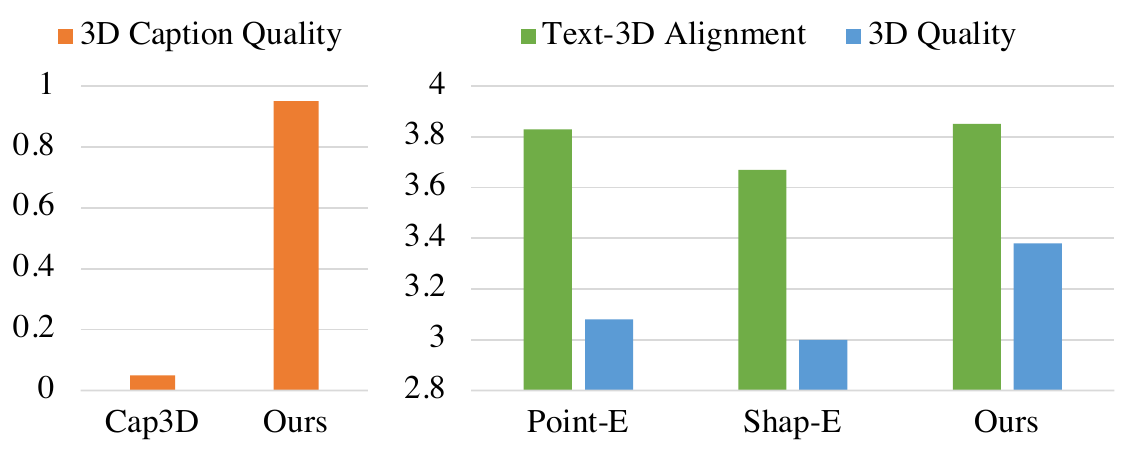}
    \caption{
    \textbf{User Study on 3D Captioning and Text-to-3D Generation}.
    The left side shows the average user preference score between Cap3D and our captions. The right chart shows the mean opinion score, ranging from 1 to 5, for text-3D alignment and 3D results quality.
    }
    \label{fig:user_study}
\end{figure}

\subsection{Quantitative Results}

In this section, we report quantitative results for different parts of our system for a better understanding of its performance. For the 3D captioning, we perform user studies and collect average user preference rates between Cap3D~\citep{cap3d} and ours, as shown in Fig.~\ref{fig:user_study}. Our caption shows a clear advantage in terms of text-3D alignment and richness.

For the tri-plane fitting, the average PSNR is \textbf{32.78}. It is a decent enough quality for the parameterization of 3D objects. Though there are better and faster 3D representations lately, the tri-plane is still the one that best combines efficiency and quality. We also compared with the non-disentangle version, which has an average PSNR of \textbf{31.48}. The disentanglement of texture and geometry helps the reconstruction quality. For the tri-plane VAE, the average PSNR is \textbf{29.56}, which is calculated between the input and reconstructed tri-plane renderings.

\begin{table}[t]
\caption{\textbf{CLIP scores of \nickname{} and comparison with Point-E and Shap-E.} We report the CLIP scores calculated with three different CLIP versions.}
\label{tab:main}
% \vskip 0.15in
\begin{center}
\begin{small}
\begin{sc}
\begin{tabular}{l|cccr}
\toprule
CLIP Version & Point-E & Shap-E & Ours \\
\midrule
laion/CLIP-ViT-bigG   & 27.08 & 30.30 & \textbf{33.28} \\
openai/CLIP-ViT-large & 18.06 & 19.73 & \textbf{22.65} \\
openai/CLIP-ViT-base  & 23.19 & 22.51 & \textbf{27.51} \\
\bottomrule
\end{tabular}
\end{sc}
\end{small}
\end{center}
\vskip -0.1in
\end{table}

\begin{figure*}[t]
    \centering
    \includegraphics[width=\textwidth]{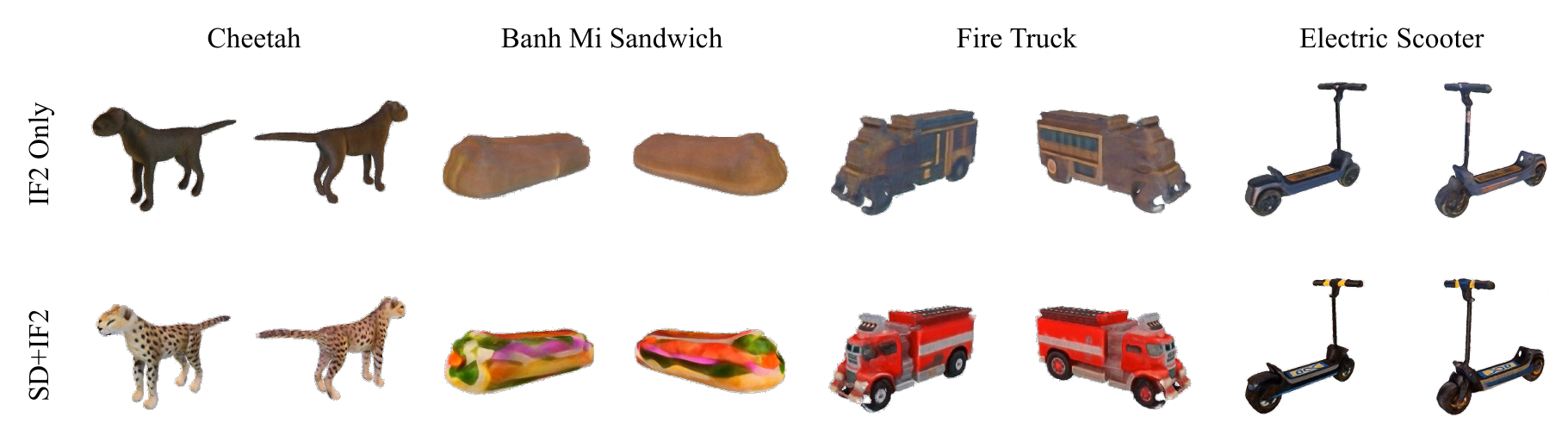}
    \caption{
    \textbf{Ablation on the necessity of latent space refinement}.
    Without applying the latent-space Stable Diffusion refinement, the texture is blurry and lacks details.
    }
    \label{fig:abl_stage2}
\end{figure*}

Next, we evaluate the whole text-to-3D pipeline and compare with two baseline methods that are open-sourced, Point-E~\citep{pointe} \footnote{\url{https://github.com/openai/point-e}} and Shap-E~\citep{shapee} \footnote{\url{https://github.com/openai/shap-e}}. Following previous text-to-3D works~\citep{pointe,shapee}, we use CLIP scores to evaluate the generation results, in terms of quality and distance to input prompts. For the testing prompts, we choose to use the prompt list proposed by~\citet{wu2023gpteval3d}, which comprises 110 prompts generated by GPT-4 to mimic user inputs. As shown in Tab.~\ref{tab:main}, we outperform both Point-E and Shap-E in CLIP scores, calculated with three different CLIP versions. Moreover, we perform user studies in terms of the text-3D alignment and 3D results quality, as shown in Fig.~\ref{fig:user_study}. For the text-3D alignment, the relative advantage of \nickname{} is small, showing that both Point-E and Shap-E are competitive baseline methods and are capable of following user inputs. \nickname{} shows larger advantages in terms of 3D results quality. Even trained with a dataset with one magnitude less than theirs, \nickname{} still shows better generation performance, thanks to our two-stage design.

\subsection{Ablation Study}

In this section, we perform ablation studies to validate design choices of \nickname{}. Firstly, for the training of latent diffusion model, we show the importance of data preparation in Tab.~\ref{tab:ablation}. Comparing the first two rows, the model trained with our own captions outperforms the one trained with Cap3D captions, indicating longer and more detailed captions are essential for high-quality text-to-3D generation. Comparing the second and third rows, the model fine-tuned with high-quality subset clearly outperforms the one without fine-tuning by a large margin, which proves the importance of data cleaning even if the high-quality subset is much smaller.

\begin{table}[t]
\caption{\textbf{Ablation study on 3D captions and data cleaning.} We report CLIP scores of \nickname{} trained with Cap3D captions or our captions. We also study the importance of fine-tuning on high-quality subset. `Cap.' stands for caption. `HQ S.S.' stands for high-quality subset.}
\label{tab:ablation}
% \vskip 0.15in
\begin{center}
\begin{small}
\begin{sc}
\begin{tabular}{ccc|c}
\toprule
% Settings & CLIP Score \\
% Only Cap3D Caption & 26.34 \\
% Only Ours Caption & 25.20 \\
% w/o High Quality Set &  \\
% w High Quality Set & 27.51 \\
Cap3D Cap. & Ours Cap. & HQ S.S. & CLIP Score \\
\midrule
\checkmark & & & 31.24 \\
& \checkmark & & 31.45 \\
& \checkmark & \checkmark & \textbf{33.28} \\
\bottomrule
\end{tabular}
\end{sc}
\end{small}
\end{center}
\vskip -0.1in
\end{table}
% Ablate on Cap3D or our caption
% Ablate on whether fine-tune with high-quality subset

% \td{} different refine priors? SD+IF2 vs IF2 -->
In Fig.~\ref{fig:abl_stage2}, we perform an ablation study on the two phases of refinement stage.
Since the pixel-space Deepfloyd IF2~\citep{deepfloydif} model is essentially a super-resolution module, we need the latent-space Stable Diffusion model to generate more details on the coarse texture from the first stage.
Otherwise, the final refined texture will look blurry and lack details.

\section{Discussion}

In this work, we propose a two-stage text-to-3D generation system, named \nickname{}. The first stage uses a text-guided latent diffusion model for quick sampling of coarse 3D models. The second stage further refines the textures of the 3D models to produce high-quality 3D assets.

\noindent\textbf{Limitation.} Admittedly, the text prompts used to generate 3D assets are simple and short. In contrast to SDS-based methods, the first stage of \nickname{} does not take advantage of any 2D priors. The amount of training data is also a few magnitudes less than those used to train text-to-2D models. Therefore, the ability to accurately generate from complex text prompts, especially concept-mixing ones, is weak. Nevertheless, in the track of feed-forward direct 3D generation, \nickname{} surpasses Point-E and Shap-E, which are trained with $10\times$ more data than we have, which shows the advantages of our design choices.

% \begin{appendices}
% \end{appendices}

\bibliographystyle{spbasic}      % basic style, author-year citations
\bibliography{ref}   % name your BibTeX data base

\end{document}